\PassOptionsToPackage{dvipsnames}{xcolor}
\documentclass[11pt,letterpaper]{style}

\usepackage[numbers]{natbib}
\usepackage{graphicx}
\usepackage{booktabs}
\usepackage{amsmath,amsfonts,amssymb}
\usepackage{subcaption}
\usepackage{multirow}
\usepackage{colortbl}
\usepackage{xparse}
\usepackage{float}
\usepackage{placeins}
\usepackage{threeparttable}
\usepackage{mathtools}
\usepackage{enumitem}
\usepackage{makecell}
\usepackage{adjustbox}
\usepackage{algorithm}
\usepackage{algorithmic}
\usepackage{caption}
\usepackage{hyperref}
\usepackage{xspace}
\graphicspath{{./}{Figures/}{figures/}{tables/}}

\hypersetup{colorlinks=true,linkcolor=red,urlcolor=blue,citecolor={blue}}

\newcommand{\methodFont}{}
\newcommand{\ours}{\methodFont{Sim2Signal}\xspace}

\renewcommand\Authfont{\centering\normalfont\bfseries\fontsize{11}{15}\selectfont}
\renewcommand\Affilfont{\centering\normalfont\fontsize{10}{15}\selectfont}

\title{Sim2Signal: Sim-to-Real Benchmarks for Traffic Signal Control}
\runningtitle{Sim2Signal: Sim-to-Real Benchmarks for Traffic Signal Control}

\author{%
    {\Authfont
    \textbf{Ferdous Al Rafi}\textsuperscript{1} \quad
    \textbf{Susrik Mukherjee}\textsuperscript{1} \quad
    \textbf{Latika Liladhar Dekate}\textsuperscript{1} \\
    \textbf{Jennifer Yawa Lavoe}\textsuperscript{1} \quad
    \textbf{Huaiyuan Yao}\textsuperscript{1} \quad
    \textbf{Shlok Mohanty}\textsuperscript{1} \\
    \textbf{Longchao Da}\textsuperscript{1} \quad
    \textbf{Xuesong Zhou}\textsuperscript{1} \quad
    \textbf{Hua Wei}\textsuperscript{1}
    }\\
    {\Affilfont
    \textsuperscript{1} Arizona State University \\
    \texttt{\{frafi1, smukhe61, ldekate, jlavoe, huaiyuan,} \\
    \texttt{smohan62, longchao, xzhou74, hua.wei\}@asu.edu}
    }
}

\begin{document}
\begin{abstract}
Reinforcement learning achieves strong traffic signal control performance in simulation, yet policies trained in simulators often fail once deployed in the real world, a failure known as the Sim-to-Real gap. When RL is applied to traffic signal control, this gap arises from several sources: sensing, action execution, traffic dynamics, and the control objective. Their relative impact and the reliability of existing Sim-to-Real mitigation methods remain insufficiently understood, and the field lacks a standard benchmark for systematically measuring the gap and evaluating mitigation methods. We present \ours, a benchmark that decomposes the Sim-to-Real gap into observation, action, transition, and reward gaps, corresponding to mismatches in the four components of the underlying MDP, and induces each gap in isolation under a shared protocol. We evaluate 18 mitigation methods on 2 base controllers, across 33 gap settings and 10 calibrated networks built from 5 real-world locations. We find that direct transfer consistently degrades performance across all four gap sources, but the severity of the degradation does not predict the effectiveness of mitigation. Instead, mitigation effectiveness depends strongly on the network and gap setting: outside the action gap, a method that helps in one case may fail in another. The most effective methods generally estimate what the gap changes, rather than make the policy insensitive through domain randomization or invariant representations. Our code is available at \url{https://github.com/DaRL-LibSignal/Sim2Signal}.
\end{abstract}

\newcommand{\TitleLinks}{%
\centering
    \vspace{8pt}
}
\maketitle

\section{Introduction}

Traffic signal control (TSC) plays a critical role in modern urban transportation systems by coordinating vehicle flows at intersections and mitigating congestion~\citep{daGenerativeAITransportation2025}. In recent years, reinforcement learning (RL) has emerged as a promising approach for adaptive traffic signal control, enabling policies to learn complex control strategies directly from data. A large body of work has demonstrated that RL-based controllers can outperform traditional rule-based or optimization-based approaches in simulated traffic environments~\citep{weiIntelliLightReinforcementLearning2018, weiRecentAdvancesReinforcement2021}. However, despite these encouraging results, deploying learned policies in real-world traffic systems remains challenging~\citep{hoferSim2RealRoboticsAutomation2021}.

One of the central obstacles is the \emph{Sim-to-Real gap}, where policies trained in simulation fail to maintain their performance when applied to real-world environments~\citep{daSim2RealTransferTraffic2023}. In traffic signal control, simulation environments inevitably simplify many aspects of real traffic systems, including sensing capabilities, action execution, traffic dynamics, and control objectives~\citep{wagenmakerOvercomingSimtoRealGap2024}. As a result, discrepancies between simulation and reality can lead to unexpected policy behaviors and significant performance degradation. Understanding the sources of these discrepancies is therefore essential for developing learning-based traffic signal controllers that are reliable in real-world deployments.

In practice, the Sim-to-Real gap in traffic signal control can arise from several distinct sources. For instance, real-world sensing systems may provide incomplete or noisy traffic observations compared with simulated states. Real traffic dynamics can differ substantially from the simplified vehicle behavior models used in simulators~\citep{SUMO2018}. Similarly, signal actions may be subject to execution delays or constraints that are not captured during training~\citep{dermanActingDelayedEnvironments2023}. Even the reward functions used for training may not reflect real-world control objectives~\citep{luDiscoveryRewardFunction2025}. Although many studies acknowledge these discrepancies, their relative effects on policy performance and the robustness of current Sim-to-Real approaches remain insufficiently understood. This is because the field \textbf{lacks a consistent, comprehensive benchmark for evaluating Sim-to-Real transfer}, a need that becomes increasingly important as learning-based traffic signal control moves toward real-world deployment.

\begin{figure}[tbp]
\centering
\includegraphics[width=\textwidth]{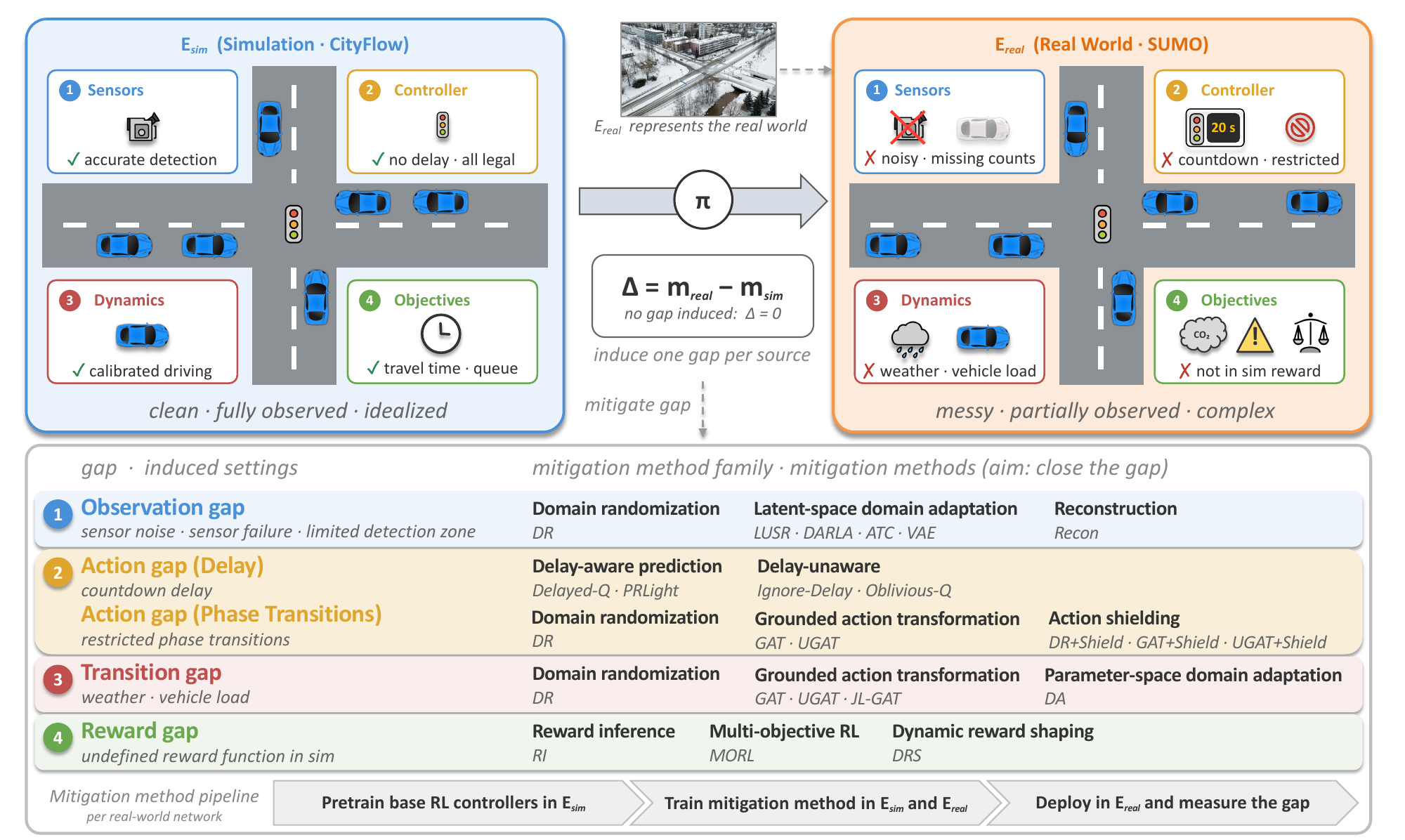}
\caption{\textbf{Overview of \ours.} A policy $\pi$ is trained in $E_{sim}$ and deployed in $E_{real}$, a second controlled environment playing the role of the real world. This sim2sim protocol isolates each of the four numbered gap sources. The middle block lists the settings induced per gap and, under each method family, all 18 mitigation methods evaluated against it. Appendix~\ref{sec:appendix_methods} expands on the method families. Every method follows the same pretrain, train, deploy pipeline under a shared episode budget, on a base controller (DQN or PressLight). Direct-Transfer applies no mitigation and is the reference. The score is $\Delta = m_{real} - m_{sim}$, with $m$ average travel time unless noted. For the reward gap, $R_{real}$ is not computable in $E_{sim}$, so methods are scored by regret against a reward oracle trained on the real objective.}
\label{fig:overview}
\end{figure}

To meet this need, we introduce a comprehensive benchmark for studying Sim-to-Real transfer in traffic signal control. Our key idea is to systematically decompose the Sim-to-Real gap into four sources, one for each component of the Markov decision process (MDP) that the environment defines: the \emph{observation}, \emph{action}, \emph{transition}, and \emph{reward} gaps. Based on this decomposition, we construct a unified experimental framework that introduces controlled discrepancies for each source, with a second controlled environment playing the role of the real world. We evaluate existing Sim-to-Real mitigation methods on standard base RL controllers to study their robustness under each gap.

We find that every gap degrades the deployed controller's performance, but that the magnitude of the degradation does not predict how much of it mitigation recovers. Recovery tends to come from estimating the quantity the gap changed, such as the state a delayed action lands in or the true vehicle count behind a noisy sensor, rather than from making the policy insensitive to the change, whether by randomized training or an invariant representation.

In summary, this work makes the following contributions:

\begin{itemize}
\item We introduce \ours, a benchmark that decomposes the Sim-to-Real problem into observation, action, transition, and reward gaps and induces each gap under a shared protocol. The benchmark covers 33 gap settings and 10 calibrated traffic networks from 5 real-world locations, including 2 new datasets. The code and datasets are included in the Code and Data Supplement.
\item We evaluate 18 Sim-to-Real mitigation methods with 2 base controllers. Running them under a shared protocol enables comparisons across gap sources, networks, settings, and base controllers.
\item  We find that the action gap has methods that work across the networks and settings we test. The other three gaps do not: a method that works on one network and setting can fail on another, and in some cases every method leaves the controller worse off than applying no mitigation.
\end{itemize}

\section{Related Work}
\label{sec:related-work}
\paragraph{RL for TSC}
Reinforcement learning has been applied to traffic signal control for two decades~\citep{abdulhaiReinforcementLearningTrue2003}. A broad family of RL controllers reports gains over traditional approaches in simulation~\citep{weiIntelliLightReinforcementLearning2018, weiPressLightLearningMax2019}, as summarized by \citet{weiRecentAdvancesReinforcement2021}. Most evaluations, however, report how well a policy performs under the conditions it trained in, not whether that performance holds once those conditions change.

\paragraph{Sim-to-Real Mitigation for TSC}
Sim-to-Real mitigation methods usually assume a known gap source and design a mechanism specifically for it. Observation-oriented work randomizes sensor inputs and studies degraded sensing, including sensor failures, partial detection, and sensor-free intersections~\citep{mullerBridgingRealityGap2023}. Action-oriented methods encode admissible phase transitions or signal-timing rules through masks, safety models, rewards, and loss constraints~\citep{duSafeLightReinforcementLearning2023}. Transition methods use grounded action transformation or meta-learning to adapt across vehicle dynamics, traffic flows, and intersections~\citep{daSim2RealTransferTraffic2023, daUncertaintyAwareGroundedAction2023, turnau2025joint, mullerBridgingRealityGap2023}. Reward-oriented studies broaden traffic objectives to include emissions, noise, pedestrians, fairness, and safety~\citep{yeFairLightFairnessAwareAutonomous2023, duSafeLightReinforcementLearning2023}, taking the deployment objective as given. These mechanisms are developed one gap at a time and evaluated on their own networks, simulators, and baselines, so which gap hurts most, and how well a method holds up beyond the setup it was built for, are both unknown.

\paragraph{Deployment-Oriented TSC}
Deployment-oriented work reduces the gap by increasing fidelity to a specific site. A recent review identifies imperfect detection, scenario mismatch, regulatory constraints, safety requirements, and the cost of online exploration as central deployment obstacles~\citep{chenRealDealReview2022}. Practical systems address them through calibrated digital twins, real demand data, state estimation from faulty detectors, and admissible signal programs~\citep{ mullerRealWorldDeploymentReinforcement2021}, or through hardware-in-the-loop execution on commercial controllers~\citep{zhangEnablingRealTimePhase2026}. These systems are built for a single site and tested end to end, so it is hard to tell which of these fixes helped, or how much of the gap they reduced.

\paragraph{TSC Benchmarks}
Existing benchmarks standardize how RL traffic signal controllers are compared, not how they transfer from simulation to deployment. Some fix a set of scenarios and baselines on a single simulator, such as RESCO~\citep{aultReinforcementLearningBenchmarks2021} and SUMO-RL~\citep{sumorl}. Others unify several simulators behind one API, so a controller can run on more than one, such as TSLib~\citep{tranTSLibUnifiedTraffic2021}, LibSignal~\citep{meiLibsignalOpenLibrary2024}, and PyTSC~\citep{bokadePyTSCUnifiedPlatform2024}. A third group moves evaluation toward deployment conditions, with a scenario built from a real intersection and its production controller logic~\citep{mullerLemgoRLOpensourceBenchmark2021} or a scaled physical endpoint~\citep{zhangReproducibleLowcostSimtoReal2025}. These provide a more realistic environment to test in, but they do not isolate the individual gap sources. None of these benchmarks evaluates mitigation methods against each gap source under a shared protocol.

Across these strands, the gap is typically identified or reduced rather than benchmarked. Some studies target one gap at a time, some build a single site end to end, and some compare controllers under idealized conditions that never change. Consequently, results reported without any change between training and evaluation say little about deployment, and the mitigation methods meant to close the Sim-to-Real gap are never themselves what gets measured. In this paper, \ours makes each gap source a controlled variable and the mitigation methods the object of the evaluation.

\section{The \ours Benchmark}
\label{sec:benchmark}

\begin{figure}[tbp]
    \centering
    \includegraphics[width=\textwidth]{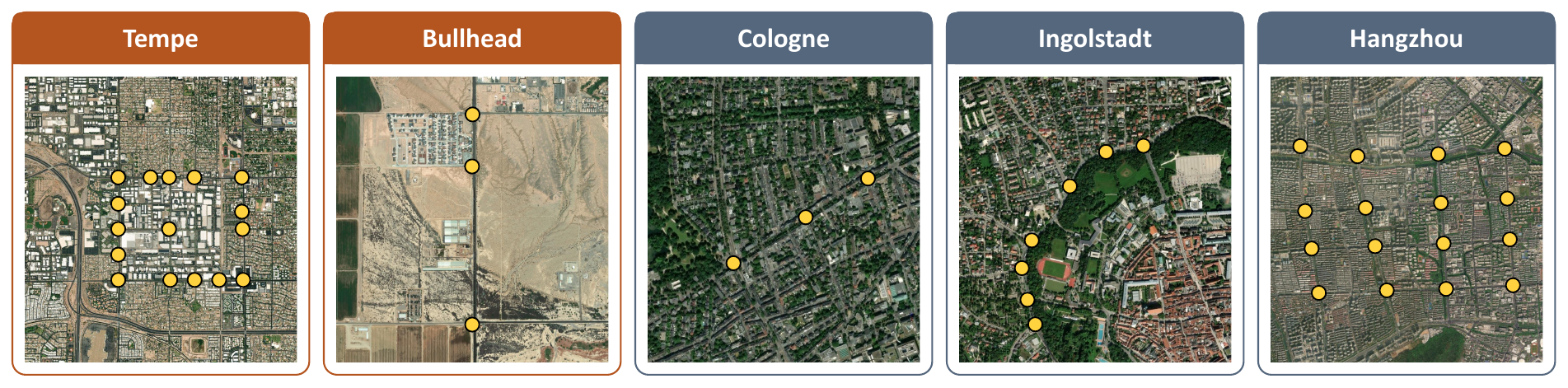}
    \caption{\textbf{The five locations of \ours.} One column per location: the real-world extent each simulated network is built from, controlled intersections marked with a yellow dot. Each location gives one single-intersection and one multi-intersection network, ten in total spanning 1 to 16 signals. The multi-intersection one is shown. All ten are calibrated to match on ATT and throughput before any gap is induced. Tempe and Bullhead (orange) are new in this work, built from real UTDF signal-plan data~\citep{luoAutomatingTrafficMicrosimulation2025}. The rest are adapted from prior research literature~\citep{aultReinforcementLearningBenchmarks2021, weiIntelliLightReinforcementLearning2018} with careful calibration. Only these two carry NEMA signal plans, so only on them can the phase-transition gap be induced.}
    \label{fig:network}
\end{figure}

\subsection{Shared Protocol}

\paragraph{Environment.}
\ours runs on LibSignal~\citep{meiLibsignalOpenLibrary2024}, which unifies CityFlow~\citep{zhangCityFlowMultiAgentReinforcement2019} and SUMO~\citep{SUMO2018}. Figure~\ref{fig:overview} gives an overview. We use the sim2sim setting~\citep{daSurveySimtoRealMethods2025}: one environment is $E_{sim}$, and a second, modified environment plays the role of $E_{real}$. Because $E_{real}$ is fully controllable, each gap can be induced in isolation and measured exactly. Policies always train in $E_{sim}$. The real environment supplies rollouts only. No method trains its policy on the real reward (the reward oracle is the sole exception). We study vehicles only and leave pedestrians to future work.

\paragraph{Pipeline.}
Every method follows the same pretrain, train, and deploy stages of Figure~\ref{fig:overview}, on a base controller of DQN~\citep{zhengDiagnosingReinforcementLearning2019} or PressLight~\citep{weiPressLightLearningMax2019}. All start from one shared pretrained checkpoint per network and controller pair, and spend at most 300 episodes, of which at most 100 run in $E_{real}$. For the observation, action, and transition gaps, deployment reports the top five evaluation checkpoints, a deliberately generous rule: a gap that survives it will not vanish under a stricter one. Reward-gap methods select on the real objective by design, so the rule does not apply to them. Appendix~\ref{sec:appendix_details} provides further details.

\paragraph{Metrics.} Average travel time (ATT) is the primary metric for the observation, action, and transition gaps, following prior Sim-to-Real work in traffic signal control~\citep{daSim2RealTransferTraffic2023}. For a metric $m$, the performance gap is $\Delta = m_{real} - m_{sim}$, so positive $\Delta$ on a cost means transfer hurts. The reward gap admits no such subtraction, because $R_{real}$ is not computable in $E_{sim}$. We instead report regret against a reward oracle, a policy trained directly in $E_{real}$ on the true objective and outside the episode budget. For a policy $\pi$, $\mathrm{regret}_{\pi} = R_{oracle} - R_{\pi}$, and $\mathrm{regret}_{DT}$ denotes the regret of Direct-Transfer. Appendix~\ref{sec:appendix_details} covers the remaining metrics and the comparability of reward regret across networks.

\subsection{Networks and Calibration}

The benchmark covers five locations: Tempe and Bullhead City (United States), Cologne and Ingolstadt (Germany), and Hangzhou (China). Each has a single-intersection and a multi-intersection network (ten total, Figure~\ref{fig:network}): Tempe 1$\times$1, Tempe 16, Bullhead 1, Bullhead 3, Cologne 1, Cologne 3, Ingolstadt 1, Ingolstadt 7, Hangzhou 1$\times$1, and Hangzhou 4$\times$4. Cologne and Ingolstadt adapt RESCO~\citep{aultReinforcementLearningBenchmarks2021} (TAPAS Cologne~\citep{varschenMikroskopischeModellierungPersonenverkehrsnachfrage2006}, InTAS~\citep{loboInTASIngolstadtTraffic2020}). Hangzhou adapts public releases from prior traffic signal control work~\citep{weiIntelliLightReinforcementLearning2018}. Tempe and Bullhead are new here: we build them from real UTDF signal-plan data~\citep{luoAutomatingTrafficMicrosimulation2025}, so they carry the NEMA dual-ring phase configurations and timings that US traffic agencies run in the field~\citep{NEMAStandardsPublication2021, urbanikSignalTimingManual2015}.

$E_{sim}$ and $E_{real}$ share network layout, demand, and signal settings. We calibrate vehicle parameters so that, with no gap induced, a policy trained in $E_{sim}$ reproduces its ATT and throughput in $E_{real}$ to within the residuals reported in Table~\ref{tab:calibration}. Every induced gap is applied after this baseline, so measured $\Delta$ is attributable to the gap rather than differences between the two simulators.

\subsection{Controlled Gap Settings}

\begin{figure}[tbp]
\centering
\includegraphics[width=\textwidth]{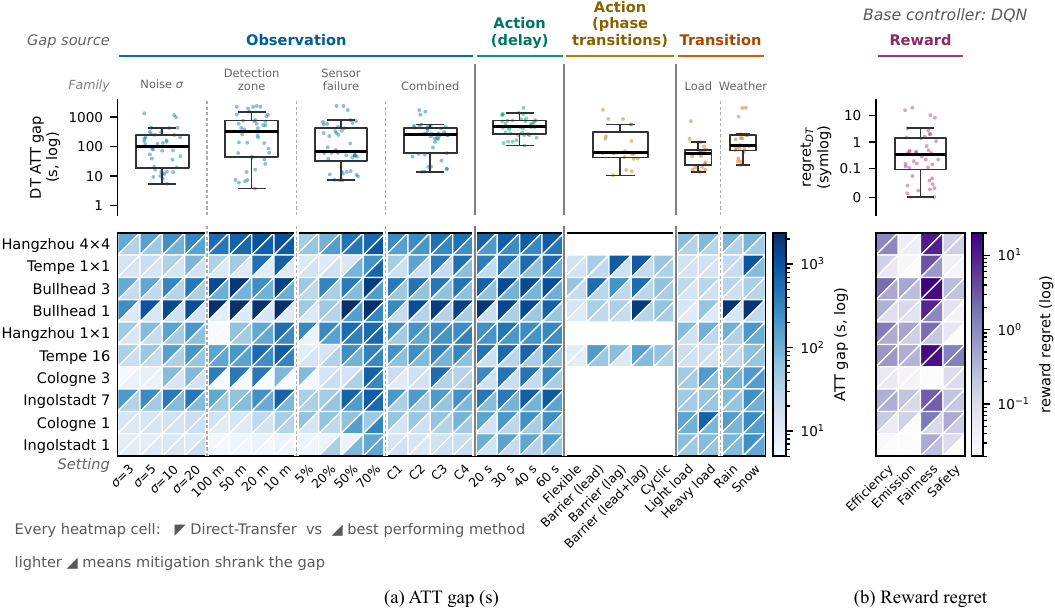}
\caption{\textbf{The Sim-to-Real gap by setting family and by network (RQ1 + RQ3).} The action delay settings have the largest gaps of any family. Gap magnitude follows congestion rather than network size. On some settings, like sensor failure, mitigation backfires and the best-method triangle comes out darker rather than lighter. Every value is one run of the DQN base controller. Figure~\ref{fig:networks_presslight} in Appendix~\ref{sec:appendix_results} repeats the figure for PressLight. \emph{Top row:} the Direct-Transfer gap distribution of each setting family, one dot per network and setting, boxes the median and interquartile range. \emph{Heatmap:} one cell per network and setting, split into the Direct-Transfer gap (upper-left triangle) and the gap under the best performing method on that cell (lower-right). The best method is picked per cell, so the lower-right triangle is an optimistic bound. Rows are ordered by travel time in simulation before any gap is induced, so the most congested network is at the top. Phase-transition cells are blank outside Tempe and Bullhead, the only networks with real NEMA signal plans. (a) ATT gap in seconds, one shared log color scale for both triangles. (b) Reward regret on its own scale. The best method is never worse than Direct-Transfer, because Reward Inference keeps the pretrained policy among its candidates. Colors compare within a cell but not across networks, since $R_{real}$ sums size-dependent terms. The full result is available in Appendix~\ref{sec:appendix_results}.}
\label{fig:networks}
\end{figure}

Across the four gaps we evaluate 18 mitigation methods, plus Direct-Transfer as the no-mitigation reference and the reward oracle as the best-case reference on the reward gap. Figure~\ref{fig:overview} shows the families per gap, and Table~\ref{tab:method_glossary} lists their members and identifiers. Appendix~\ref{sec:appendix_methods} provides the implementation details.

\paragraph{Observation.} Real agents read sensors, not the full state: $o = h(s)$ rather than $s$. We corrupt $E_{real}$ observations along three axes: Gaussian noise with $\sigma \in \{3, 5, 10, 20\}$~\citep{aslaniAdaptiveTrafficSignal2017}, per-lane detector failure with probability $p \in \{0.05, 0.2, 0.5, 0.7\}$ held for the episode, and detection zones of 10, 20, 50, and 100\,m from the stop line. Combined settings C1--C4 stack these sources as specified in Table~\ref{tab:obs_combine}.

\textit{Mitigation methods.} Domain randomization, latent-space domain adaptation (LUSR~\citep{xingDomainAdaptationReinforcement2021}, DARLA~\citep{higginsDARLAImprovingZeroShot2018}, ATC~\citep{stookeDecouplingRepresentationLearning2021}, and a plain VAE embedding~\citep{kingmaAutoEncodingVariationalBayes2022}), and reconstruction. Domain adaptation here means aligning observation distributions between $E_{sim}$ and $E_{real}$. Each latent encoder is pretrained on domain-randomized observations and frozen, while reconstruction feeds the policy a denoised observation rather than a latent code. Domain randomization draws a fresh joint corruption per episode, and the two harshest noise and failure levels fall outside the range it trains on.

\paragraph{Action (delay).} Real controllers often apply a countdown before a requested phase takes effect~\citep{panImpactCountdownSignals2023}. We model a constant-delay MDP~\citep{dermanActingDelayedEnvironments2023}: decisions every 10\,s, with $E_{real}$ delays of 20, 30, 40, and 60\,s. Mitigation methods assume a 20\,s countdown, so larger delays also test underestimation of the delay.

\textit{Mitigation methods.} Delay-aware prediction, covering Delayed-Q~\citep{dermanActingDelayedEnvironments2023}, which rolls a forward model through the pending actions, and PRLight~\citep{hanMitigatingActionHysteresis2023}, which predicts post-countdown features in one step. Two ablations isolate what prediction adds: Ignore-Delay trains delay-free on the same simulated budget, and Oblivious-Q~\citep{dermanActingDelayedEnvironments2023} trains under the delay but still acts on the current observation.

\paragraph{Action (phase transitions).} Simulators allow any phase successor, while real programs forbid many switches via minimum green, clearance, and fixed sequences~\citep{urbanikSignalTimingManual2015}. On Tempe and Bullhead we enforce NEMA dual-ring rules in $E_{real}$ only, at five settings: Flexible, Barrier (lead), Barrier (lag), Barrier (lead+lag), and Cyclic. Flexible forbids no transitions but keeps the minimum and maximum green times. The barrier variants disable roughly 43\% of transitions, and Cyclic fixes one successor per phase. A disallowed request is dropped and the current phase held.

\textit{Mitigation methods.} Domain randomization over random rule tables and grounded action transformation (GAT~\citep{hannaGroundedActionTransformation2021}, UGAT~\citep{daUncertaintyAwareGroundedAction2023}), each with and without deployment-time action shielding~\citep{alshiekhSafeReinforcementLearning2018}, which masks the action set to the currently legal transitions and substitutes the highest-ranked allowed action.

\paragraph{Transition.} Dynamics differ when driver behavior, vehicle loading, or weather changes. CityFlow is $E_{sim}$ and SUMO is $E_{real}$. We further perturb Krauss car-following parameters in $E_{real}$~\citep{kraussMicroscopicModelingTraffic1998} into four settings (Light load, Heavy load, Rain, Snow) using the published calibrations listed in Table~\ref{tab:transition_settings}~\citep{pourabdollahCalibrationEvaluationCar2017, schraderCalibratingCarFollowingModels2024}.

\textit{Mitigation methods.} Domain randomization~\citep{tobinDomainRandomizationTransferring2017a, mullerBridgingRealityGap2023}, parameter-space domain adaptation via simulation-based inference~\citep{ramosBayesSimAdaptiveDomain2019a}, and grounded action transformation (GAT~\citep{hannaGroundedActionTransformation2021}, UGAT~\citep{daUncertaintyAwareGroundedAction2023}, JL-GAT~\citep{turnau2025joint} on multi-intersection networks). Domain randomization uses no real rollouts, while adaptation and grounding each run 100 episodes that pair one simulated rollout with one real one.

\begin{figure}[tbp]
\centering
\includegraphics[width=\textwidth]{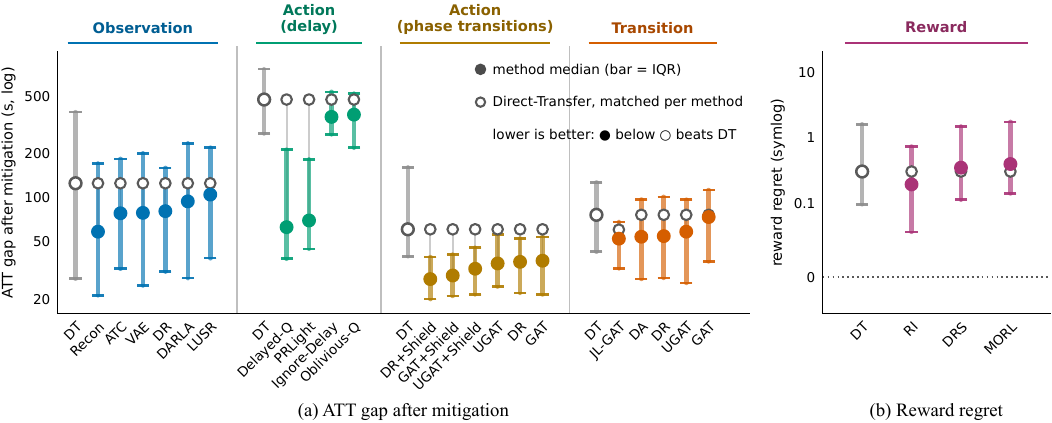}
\caption{\textbf{Recovery under mitigation (RQ2).} Mitigation recovers most of the two action gaps and less of the observation, transition, and reward gaps, and is sometimes worse than Direct-Transfer. The methods that recover most tend to be the ones that estimate what the gap changed. Delay, the most severe gap, also recovers the most, so severity does not predict recovery. A bar crossing its open circle marks a method that helps on some networks and settings and hurts on others. One column per method, grouped by gap source and sorted by median within a group. Short forms follow Figure~\ref{fig:overview}. Each group leads with a grey Direct-Transfer column pooled over that gap's deployments. Per column, the filled circle is the median and the bar the IQR. The open circle is the matched Direct-Transfer median, so it shifts from column to column. (a) ATT gap after mitigation, all four gap sources on one shared log scale. (b) Reward regret, $\mathrm{regret}_{\pi} = R_{oracle} - R_{\pi}$, on a symmetric log scale in its own units. The oracle is the reference, at regret 0 by definition (dotted line).}
\label{fig:recovery}
\end{figure}

\paragraph{Reward.} Agencies weigh delay against emissions, safety, and fairness~\citep{urbanikSignalTimingManual2015}, while simulators expose only a subset of these terms. Training uses the controller's native reward, and deployment scores a hidden objective $R_{real}$. Both are weighted sums of normalized cost components, $R = -\sum_i w_i \phi_i$. The hidden objective reweights the shared components and adds components $\phi_j^{real}$ computable only in $E_{real}$ (CO$_2$, time-to-collision conflicts)~\citep{mahmudApplicationProximalSurrogate2017}. Four settings reweight the objective, one toward efficiency and one each toward emissions, safety, and fairness, the last two built on the real-only components. Table~\ref{tab:reward_settings} gives their weights. Methods observe the real objective only as a per-episode return from deployed rollouts, never as a per-step reward.

\textit{Mitigation methods.} No existing method targets the reward gap, so we construct three from standard components. Reward inference recovers the hidden weights by regression on probe rollouts and retrains in simulation. Multi-objective RL trains a grid of simulator-visible scalarizations. Dynamic reward shaping~\citep{muratoreDataefficientDomainRandomization2021} runs a Bayesian search over the weights. All three deploy the candidate with the best real return and are scored against the reward oracle. The three differ in how they use the pretrained policy. By design, reward inference is the conservative one. It alone keeps the pretrained policy among its candidates, so it never deploys below Direct-Transfer. The other two pick the best of what they train, and that can be worse.

\section{Benchmarking Results}
\label{sec:results}


We organize the evaluation around four research questions:

\noindent$\bullet$~\textbf{RQ1 (gap severity):} How severely does each gap source degrade Direct-Transfer?

\noindent$\bullet$~\textbf{RQ2 (mitigation effectiveness):} How much of each gap can existing mitigation methods recover?

\noindent$\bullet$~\textbf{RQ3 (network conditions):} How do gap severity and mitigation effectiveness vary across network configurations?

\noindent$\bullet$~\textbf{RQ4 (controller robustness):} Do these findings hold across base controllers?

Queue and throughput follow the same method rankings as ATT, so the main analysis reports ATT. Appendix~\ref{sec:appendix_results} gives all metrics and per-setting results.

\subsection{Gap Severity by Source (RQ1)}

Before comparing mitigation methods, we need to know how much each gap source degrades the deployed controller's performance on its own, since that degradation is what a method has to recover. We deploy Direct-Transfer into every induced setting on every network and record the ATT gap $\Delta$, reading severity as the size of that gap and as regret for the reward gap. The distribution row along the top of Figure~\ref{fig:networks} ranks the gap sources and the setting families within them for DQN. Figure~\ref{fig:networks_presslight} repeats it for PressLight, and Appendix~\ref{sec:appendix_results} gives the full per-network and per-setting values.

Performance drops under every gap source, and by very different amounts. The action delay gap is the most severe, and no other setting family has a higher median. Restricted phase transitions are among the lowest, and the observation and transition gaps fall in between. Within the observation gap, severity varies by corruption type, with sensor failure producing the smallest median gap and the reduced detection zone the largest. Reward regret remains positive across all networks and settings, indicating that Direct-Transfer consistently underperforms the oracle. Overall, severity varies substantially both within and across gap sources, so the gap category alone does not determine its impact.

\subsection{Mitigation Effectiveness Across Gaps (RQ2)}

Since every gap source can severely degrade the deployed controller's performance, the next question is how much of that degradation can be mitigated, and which approaches work. On every network and setting where a gap is induced, we run the mitigation methods built for it and compare each against Direct-Transfer, aggregating over both base controllers. Figure~\ref{fig:recovery} reports the result, and Appendix~\ref{sec:appendix_results} gives the per-method and per-setting values.

\paragraph{Both action gaps shrink reliably.} On the delay gap, Delayed-Q and PRLight, which predict the state a delayed action lands in, shrink the gap sharply. Ignore-Delay and Oblivious-Q shrink it far less. Oblivious-Q trains under the same delay and still recovers little, so predicting the delayed state, rather than merely training under the delay, is what removes most of the gap. That holds even though every method assumes a 20\,s countdown against real delays of up to 60\,s. On the phase transition gap, domain randomization and the two grounding methods all improve on Direct-Transfer, and adding action shielding lowers each of them further. Both action gaps are therefore largely recoverable, though by different means. Delay is the most severe gap and phase transitions among the least severe, so severity does not tell us how much of a gap we can recover.

\paragraph{The observation, transition, and reward gaps shrink unreliably.} Most methods help on average, but not reliably. The same method can shrink the gap in one deployment and make it worse in another. On the observation gap the reconstruction baseline, which estimates the true reading behind the corrupted one, has the lowest median of any method, ahead of every latent-space method, which instead learns a representation that ignores the corruption. On the transition gap four of the five methods recover about the same amount, and GAT recovers almost nothing. On the reward gap Reward Inference, the only method that estimates the hidden weights, is the only one that improves on Direct-Transfer. The other two end worse than applying no mitigation at all. The DQN heatmap in Figure~\ref{fig:networks} breaks the same results out by network and setting. On the sensor-failure rows the best method comes out darker than Direct-Transfer, so the setting with the smallest observation gap is the one mitigation makes worse. On these three gaps, therefore, no method gives an improvement that holds on every network and setting.

Across all four gap sources, the methods that recover most tend to be the ones that estimate what the gap changed, such as the state a delayed action lands in (Delayed-Q, PRLight) or the true reading behind a noisy sensor (Reconstruction). On the observation gap, the latent-space methods learn to ignore the change instead, and they recover less. On the transition gap, estimating the changed dynamics parameters (parameter-space domain adaptation) and training across a range of them (domain randomization) recover about the same amount. On the phase transition gap, shielded domain randomization recovers the most, with shielded GAT close behind. The pattern therefore does not hold everywhere. It appears often enough to be worth investigating further.

\subsection{Variation Across Network Conditions (RQ3)}

So far we have compared gap sources and mitigation methods, but not how either varies with the network the controller is deployed on. We hold the setting fixed and vary the deployment network, which isolates what the network itself contributes. Each network is one row of the DQN ATT heatmap in Figure~\ref{fig:networks}, where a darker cell means a larger gap. Rows are ordered by travel time in simulation before any gap is induced, so the most congested network is at the top. Appendix~\ref{sec:appendix_results} gives the per-network values.

The largest gaps tend to be on congested networks. Bullhead 1 and Bullhead 3 have the largest of any network, and Ingolstadt 1, the least congested, has the smallest by a wide margin. The ordering is rough, since Tempe 1$\times$1 is the second most congested network and among the smallest. Mitigation also recovers the most on the networks with the largest gaps. On Ingolstadt 1 the best method does not improve on Direct-Transfer, and several settings end worse after mitigation than before. Congested networks therefore lose the most and recover the most. On the least congested network there is little to recover, and mitigation can do more harm than good.

\begin{figure}[tbp]
\centering
\includegraphics[width=0.62\textwidth]{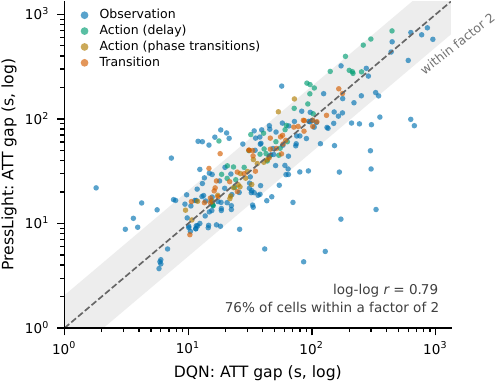}
\caption{\textbf{Consistency across base controllers (RQ4).} The findings largely do not depend on the base controller the mitigation is applied to. Most networks and settings agree within a factor of two. Each dot is one network and setting after mitigation, under the best performing method there, with DQN on $x$ and PressLight on $y$, colored by gap source. The shaded band is agreement within a factor of two ($0.5 \le y/x \le 2$), and $r$ is the Pearson correlation of the $\log_{10}$ values. Figure~\ref{fig:controllers_reward} in Appendix~\ref{sec:appendix_results} is the reward-regret counterpart.}
\label{fig:controllers}
\end{figure}
\subsection{Robustness Across Base Controllers (RQ4)}

Every result so far was measured on one base controller, so a verdict about a gap could belong to that controller rather than to the gap itself. We repeat the benchmark with PressLight in place of DQN and compare the two on each network and setting, under the best mitigation method there. Figure~\ref{fig:controllers} plots the ATT gap and Figure~\ref{fig:controllers_reward} the reward counterpart, and Appendix~\ref{sec:appendix_results} gives the per-controller values.

The two controllers agree on 76\% of networks and settings for the ATT gap, at $r=0.79$, and on 70\% for reward regret, at $r=0.91$. Agreement holds across all four gap sources. Most of the disagreements beyond a factor of two belong to the observation gap. Some of those are worse under DQN and some under PressLight, so neither controller is more robust than the other. The method that recovers the most also stays the same under both controllers on the observation, delay, and phase transition gaps (Appendix~\ref{sec:appendix_results}). The gap left after mitigation is therefore mostly a property of the network and setting rather than of the controller it was measured on.

\section{Conclusion}
\label{sec:conclusion}


In this work, we introduced \ours as a controlled benchmark for isolating and evaluating the observation, action, transition, and reward gaps in Sim-to-Real traffic signal control. The results show that deployment degradation depends on the gap configuration and network conditions, rather than on the gap category alone, and that gap severity is not a reliable indicator of mitigation effectiveness. Methods that explicitly estimate the altered state, dynamics, action execution, or objective generally achieve more consistent recovery than approaches based primarily on domain randomization or invariant representations. These findings motivate mitigation methods that identify and correct the specific deployment gap instead of relying on generic robustness mechanisms. \ours provides a standardized testbed for evaluating whether such methods generalize across networks, settings, and base controllers. Future extensions will include additional gap sources, such as lane closures and pedestrian interactions, as well as settings in which multiple gaps occur simultaneously.

\bibliographystyle{unsrtnat}
\bibliography{references}

\appendix
\newcommand{\mainsecresults}{Section~\ref{sec:results}}
\newcommand{\mainfignetworks}{Figure~\ref{fig:networks}}
\newcommand{\mainfigcontrollers}{Figure~\ref{fig:controllers}}

\setcounter{table}{0}
\setcounter{figure}{0}
\renewcommand{\thetable}{A\arabic{table}}
\renewcommand{\thefigure}{A\arabic{figure}}

\let\sstabular\tabular
\let\endsstabular\endtabular
\renewenvironment{tabular}%
  {\begin{adjustbox}{max width=\linewidth}\sstabular}%
  {\endsstabular\end{adjustbox}}

\section{Benchmark Details}
\label{sec:appendix_details}

\paragraph{Scale.} Table~\ref{tab:benchmark_scale} breaks the benchmark down: networks, induced settings, and methods per gap, and the resulting number of deployment runs. Every count quoted in the main text can be recovered from it. Table~\ref{tab:method_glossary} maps every mitigation method to the identifier it carries in the results tables.

\paragraph{Calibration.} Table~\ref{tab:calibration} reports the average travel time of the shared pretrained checkpoint in both environments under the nominal setting, with no gap induced. The no-gap residual is 0.3--19\,s across all ten networks and both base controllers, median 9\,s and 7\,s for DQN alone, against induced gaps in \mainsecresults{} that reach hundreds to thousands of seconds. Transfer degradation therefore comes mainly from the induced gap, not from the simulator pair or interface.

\paragraph{Metrics.} Besides ATT we report throughput, average queue length, and average delay, and the reward and phase-transition settings add gap-specific metrics. Two regret quantities recur. $\mathrm{regret}_{DT}$, the regret of Direct-Transfer, measures the real objective given up by a simulator-trained policy that never observes the deployment objective. $\mathrm{regret}_{best}$ is the regret of the best budgeted method in a deployment. Because $R_{real}$ sums per-intersection costs under shared normalization, its scale grows with network size and demand. Reward regret is therefore comparable across methods within a deployment, but only ordinally across networks. Aggregate results are taken over networks, settings, and both base controllers, and Appendix~\ref{sec:appendix_results} reports all metrics with full per-setting results.

\paragraph{Protocol.} Each episode covers 3600 steps, one hour of traffic. Pretraining runs 200 episodes per network and base controller pair under the nominal setting with default LibSignal hyperparameters~\citep{meiLibsignalOpenLibrary2024}, and the checkpoint is selected to match in $E_{sim}$ and $E_{real}$ when no gap is induced. Every mitigation method begins training from that shared checkpoint, so comparisons are not confounded by pretraining variance, and the checkpoints are released with the benchmark. Deployment evaluates for up to 100 episodes in $E_{real}$, and the reported mean $\pm$ std is taken over the top five evaluation checkpoints. Reward-gap methods select on the real objective by design, so the top-five rule does not apply to them, and the reward oracle is a reference rather than a budgeted method.

\paragraph{Observation gap.} Table~\ref{tab:obs_combine} defines the combined-corruption settings. The single-source settings use the values given in the main text.

\paragraph{Transition gap.} Table~\ref{tab:transition_settings} lists the car-following parameters of the four settings. Table~\ref{tab:transition_dr} gives the sampling distributions used by domain randomization.

\paragraph{Reward gap.} Table~\ref{tab:reward_settings} lists the hidden-objective weights per setting.


\paragraph{Methods.} Table~\ref{tab:method_glossary} expands the method families named in the overview figure of the main text into the individual methods evaluated under each, and gives the identifier each carries in the full-results tables. The Method Details subsection below records implementation choices that the main text compresses to one-line family descriptions.

\paragraph{Seeding and infrastructure.} Every run reported here uses seed 0, set in the shipped configuration files. The seed is applied to Python, NumPy, and PyTorch, including the CUDA generators, and the cuDNN backend is put in deterministic mode. A command-line flag overrides it for anyone who wants to repeat a run under a different seed. Every hyperparameter of every method is fixed in the shipped configuration files, and we release the pretrained checkpoints alongside them. Deployment always runs in SUMO, which is deterministic, so evaluating a released checkpoint returns exactly the values reported here. Training runs in CityFlow, which steps the simulation across several threads and varies slightly between runs even at a fixed seed. A policy retrained from scratch therefore lands close to the reported value rather than on it. Experiments were run on (i) a workstation with an Intel Core Ultra 9 285K (24 cores), 64\,GB RAM, and an NVIDIA RTX 5090 (32\,GB), running Ubuntu 24.04, and (ii) CPU nodes of an institutional HPC cluster (dual AMD EPYC 7713, 128 cores, 512\,GB RAM per node), where each job used 4 cores and 8\,GB RAM; no GPUs were used on the cluster. Software: Python 3.10, PyTorch 2.10, SUMO 1.26.0 (libsumo), CityFlow 0.1. Neither machine is required to reproduce a result. The benchmark runs on CPU by default, so a single Ubuntu desktop reproduces any run in the tables, given enough wall-clock time. The code supplement pins every Python dependency with its version.

\begin{table}[h]
\centering
\small
\setlength{\tabcolsep}{4pt}
\caption{\textbf{Benchmark scale.} Every count claimed in the main paper is reconstructible from this table. Runs per gap are networks $\times$ methods $\times$ settings $\times$ 2 base controllers, and each run is one row of the full-results tables in Section~\ref{sec:appendix_results}. The methods column counts everything deployed against that gap, including Direct-Transfer, which applies no mitigation and is the reference, and, for the reward gap, the oracle. Summing that column double-counts the six mitigation methods that are evaluated against more than one gap, so the total row instead gives the 18 distinct mitigation methods, which Table~\ref{tab:method_glossary} lists individually. The transition gap is the one row that is not a plain product: JL-GAT applies only to the five multi-intersection networks, contributing $5 \times 1 \times 4 \times 2 = 40$ runs on top of the $10 \times 5 \times 4 \times 2 = 400$ from the other five methods. The phase-transition setting requires real NEMA signal plans, which only the four Tempe and Bullhead networks carry.}
\label{tab:benchmark_scale}
\begin{tabular}{@{}lcccc@{}}
\toprule
Gap & Networks & Methods & Settings & Runs \\
\midrule
Observation & 10 & 7 & 16 & 2{,}240 \\
Action (delay) & 10 & 5 & 4 & 400 \\
Action (phase transitions) & 4 & 7 & 5 & 280 \\
Transition & 10 & 6 & 4 & 440 \\
Reward & 10 & 5 & 4 & 400 \\
\midrule
Total & 10 & 18 & 33 & 3{,}760 \\
\bottomrule
\end{tabular}
\end{table}

\begin{table*}[tbp]
\centering
\small
\setlength{\tabcolsep}{4pt}
\caption{\textbf{Method families.} Expansion of the family names used in the overview figure of the main paper. Gap numbers match that figure: 1 observation, 2 action, 3 transition, 4 reward. The identifier column gives the name each method carries in the full-results tables of Section~\ref{sec:appendix_results}. Action shielding is a deployment-time choice rather than a parallel family: it combines with any adaptation method for the restricted-transition setting, so each of those methods is reported both shielded and unshielded. The final block lists the two references, which are not mitigation methods and are excluded from the count of 18: Direct-Transfer applies no mitigation, and the reward oracle is trained directly on the real objective. Ignore-Delay and Oblivious-Q are counted among the 18, since both are strategies a practitioner could actually deploy against the delay gap. They are marked as ablations because their role in the analysis is to isolate what training under the delay buys without state prediction.}
\label{tab:method_glossary}
\begin{tabular}{@{}p{3.1cm}cp{6.3cm}p{4.6cm}@{}}
\toprule
Family & Gap & Methods evaluated & Identifier in results tables \\
\midrule
Domain randomization & 1, 2, 3 & Per-episode randomization over the induced perturbation distribution~\citep{tobinDomainRandomizationTransferring2017a, mullerBridgingRealityGap2023, schraderCalibratingCarFollowingModels2024} & \texttt{domain\_randomization}, \texttt{dr}, \texttt{dr\_noshield} \\
\addlinespace[2pt]
Latent-space domain adaptation & 1 & LUSR~\citep{xingDomainAdaptationReinforcement2021}, DARLA~\citep{higginsDARLAImprovingZeroShot2018}, ATC~\citep{stookeDecouplingRepresentationLearning2021}, plain VAE embedding~\citep{kingmaAutoEncodingVariationalBayes2022} & \texttt{lusr}, \texttt{darla}, \texttt{atc}, \texttt{vae} \\
\addlinespace[2pt]
Reconstruction & 1 & Autoencoder feeding the policy its denoised reconstruction rather than a latent code & \texttt{recon\_baseline} \\
\addlinespace[2pt]
Delay-aware prediction & 2 & Delayed-Q~\citep{dermanActingDelayedEnvironments2023}, PRLight~\citep{hanMitigatingActionHysteresis2023} & \texttt{delayed\_q}, \texttt{prlight} \\
\addlinespace[2pt]
Grounded action transformation & 2, 3 & GAT~\citep{hannaGroundedActionTransformation2021}, UGAT~\citep{daUncertaintyAwareGroundedAction2023}, JL-GAT~\citep{turnau2025joint} on the multi-intersection networks only & \texttt{gat}, \texttt{ugat}, \texttt{jlgat} \\
\addlinespace[2pt]
Action shielding & 2 & Deployment-time masking of the action set to the currently legal transitions~\citep{alshiekhSafeReinforcementLearning2018} & \texttt{gat\_shield}, \texttt{ugat\_shield}, \texttt{dr} \\
\addlinespace[2pt]
Parameter-space domain adaptation & 3 & Simulation-based inference over the five car-following parameters~\citep{ramosBayesSimAdaptiveDomain2019a} & \texttt{domain\_adaptation} \\
\addlinespace[2pt]
Reward inference & 4 & Ridge regression of the real weights $w^{real}$ from probe-policy rollouts & \texttt{reward\_inference} \\
\addlinespace[2pt]
Multi-objective RL & 4 & Vector reward scalarized over a fixed grid of weights, selected by real evaluation & \texttt{morl\_grid} \\
\addlinespace[2pt]
Dynamic reward shaping & 4 & Bayesian-optimization weight search with active reward learning~\citep{muratoreDataefficientDomainRandomization2021} & \texttt{dynamic\_reward\_shaping} \\
\midrule
Direct-Transfer \emph{(reference)} & 1--4 & Pretrained policy deployed with no mitigation applied & \texttt{direct\_transfer} \\
\addlinespace[2pt]
Reward oracle \emph{(reference)} & 4 & Trained directly on the real objective; upper bound for the reward gap & \texttt{reward\_oracle} \\
\addlinespace[2pt]
Ignore-Delay \emph{(ablation)} & 2 & Trains in a delay-free simulator on the same budget and acts as if no delay exists & \texttt{naive} \\
\addlinespace[2pt]
Oblivious-Q \emph{(ablation)} & 2 & Trains under the delay but acts on the current observation~\citep{dermanActingDelayedEnvironments2023} & \texttt{oblivious\_q} \\
\bottomrule
\end{tabular}
\end{table*}

\begin{table}[t]
\centering
\scriptsize
\setlength{\tabcolsep}{3.5pt}
\caption{\textbf{Calibration.} ATT (s) of the shared pretrained checkpoint in the calibrated $E_{sim}$ (CityFlow) and $E_{real}$ (SUMO) under the nominal setting, and the residual simulator gap.}
\label{tab:calibration}
\begin{tabular}{lrrrrrr}
\toprule
 & \multicolumn{3}{c}{DQN} & \multicolumn{3}{c}{PressLight} \\
\cmidrule(lr){2-4}\cmidrule(lr){5-7}
Network & $E_{sim}$ & $E_{real}$ & gap & $E_{sim}$ & $E_{real}$ & gap \\
\midrule
tempe\_1x1 & 143.0 & 151.2 & 8.2 & 143.1 & 159.4 & 16.3 \\
bullhead\_1 & 117.9 & 127.3 & 9.4 & 121.0 & 129.1 & 8.1 \\
cologne1 & 42.1 & 55.4 & 13.3 & 44.0 & 56.8 & 12.8 \\
ingolstadt1 & 30.2 & 36.0 & 5.8 & 32.5 & 40.2 & 7.7 \\
hz1x1 & 113.8 & 114.1 & 0.3 & 112.2 & 120.5 & 8.3 \\
tempe\_16 & 76.1 & 81.9 & 5.8 & 76.2 & 91.7 & 15.5 \\
bullhead\_3 & 125.3 & 144.6 & 19.3 & 128.1 & 147.0 & 18.9 \\
cologne3 & 67.4 & 71.0 & 3.6 & 63.9 & 71.2 & 7.3 \\
ingolstadt7 & 63.7 & 82.8 & 19.1 & 63.9 & 79.3 & 15.4 \\
hz4x4 & 355.1 & 358.9 & 3.8 & 349.2 & 366.9 & 17.7 \\
\bottomrule
\end{tabular}
\end{table}

\begin{table}[htbp]
\centering
\scriptsize
\setlength{\tabcolsep}{4pt}
\caption{\textbf{Transition gap settings.} SUMO Krauss car-following parameters of $E_{real}$ per setting, following the perturbations of \citet{daUncertaintyAwareGroundedAction2023}. The maximum speed is 13.39\,m/s throughout. Vehicle length is 4.3\,m nominal and the SUMO default of 5.0\,m in all four settings.}
\label{tab:transition_settings}
\begin{tabular}{lccccc}
\toprule
Parameter & Nominal & Light & Heavy & Rain & Snow \\
\midrule
Min.\ gap (m)            & 0.5 & 2.5  & 2.5  & 2.5  & 2.5 \\
Headway $\tau$ (s)       & 1.0 & 1.5  & 1.5  & 1.5  & 1.5 \\
Accel.\ (m/s$^2$)        & 2.6 & 1.0  & 1.0  & 0.75 & 0.5 \\
Decel.\ (m/s$^2$)        & 4.5 & 2.5  & 2.5  & 3.5  & 1.5 \\
Emerg.\ decel.\ (m/s$^2$)& 9.0 & 6.0  & 6.0  & 6.0  & 2.0 \\
Startup delay (s)        & 0.0 & 0.5  & 0.75 & 0.25 & 0.5 \\
\bottomrule
\end{tabular}
\end{table}

\begin{table}[htbp]
\centering
\scriptsize
\setlength{\tabcolsep}{4pt}
\caption{\textbf{Transition gap domain randomization.} Per-episode sampling distributions over the CityFlow car-following parameters: independent Gaussians, clipped to the given range. The distributions combine the randomization of \citet{mullerBridgingRealityGap2023} with the calibrated parameter values of \citet{schraderCalibratingCarFollowingModels2024}.}
\label{tab:transition_dr}
\begin{tabular}{lccc}
\toprule
Parameter & Mean & Std & Clip range \\
\midrule
Headway time (s)          & 2.10 & 1.77 & $[0.5, 5.0]$ \\
Min.\ gap (m)             & 2.90 & 0.36 & $[1.5, 4.0]$ \\
Accel.\ (m/s$^2$)         & 2.63 & 1.36 & $[0.1, 7.0]$ \\
Decel.\ (m/s$^2$)         & 3.70 & 1.56 & $[1.0, 7.0]$ \\
Vehicle length (m)        & 5.00 & 0.64 & $[4.7, 5.0]$ \\
\bottomrule
\end{tabular}
\end{table}

\begin{table}[htbp]
\centering
\scriptsize
\setlength{\tabcolsep}{4pt}
\caption{\textbf{Combined observation corruptions.} Each combined setting applies all three corruption sources at once.}
\label{tab:obs_combine}
\begin{tabular}{lccc}
\toprule
Setting & Detection zone (m) & Noise $\sigma$ & Failure prob.\ $p$ \\
\midrule
C1 & 100 & 3 & 0.05 \\
C2 & 50  & 3 & 0.10 \\
C3 & 20  & 3 & 0.20 \\
C4 & 10  & 5 & 0.25 \\
\bottomrule
\end{tabular}
\end{table}

\begin{table}[htbp]
\centering
\scriptsize
\setlength{\tabcolsep}{3pt}
\caption{\textbf{Reward gap settings.} Component weights of the hidden deployment objective $R_{real} = -\sum_i w_i\, \phi_i$ per setting, where blank means weight zero. Components are normalized costs. Emission and TTC conflicts (TTC $<$ 1.5\,s) are computable in SUMO only, so the simulator-trained policy cannot observe them.}
\label{tab:reward_settings}
\begin{tabular}{lcccccc}
\toprule
Setting & Queue & Delay & Waiting & Emission & Fairness & TTC Confl. \\
\midrule
Efficiency-aligned & 1.0 & 1.0 & 0.5 &     &     &     \\
Emission-heavy     & 0.5 &     &     & 2.0 &     &     \\
Fairness-heavy     & 0.5 &     & 1.0 &     & 2.0 &     \\
Safety-heavy       & 0.5 &     &     &     &     & 2.0 \\
\bottomrule
\end{tabular}
\end{table}

\subsection{Benchmark Formulation}
\label{sec:appendix_formulation}

We model traffic signal control as a Markov Decision Process (MDP). At time $t$, a policy $\pi$ maps an observation $o_t$ to a requested signal action $a_t$. The environment executes an action $\tilde{a}_t$, transitions from state $s_t$ to $s_{t+1}$ according to $T(s_{t+1}\mid s_t,\tilde{a}_t)$, and evaluates the outcome with reward $R(s_t,\tilde{a}_t,s_{t+1})$. This distinction between the requested and executed action allows the formulation to represent deployment delays and signal-transition constraints.
In a Sim-to-Real setting~\citep{daSurveySimtoRealMethods2025}, we distinguish a training environment $E_{sim}$ from a deployment environment $E_{real}$. A policy is trained using interactions and rewards available in $E_{sim}$, then evaluated in $E_{real}$. Sim-to-Real transfer fails when a gap between the two environments degrades deployment performance.

\paragraph{Gap decomposition.}

\ours decomposes this gap along four parts of the control loop. An \emph{observation gap} changes how the underlying state is measured: $o_t^{sim}=h_{sim}(s_t)$ and $o_t^{real}=h_{real}(s_t)$. An \emph{action gap} changes how a requested action is executed: $\tilde{a}_t^{sim}=g_{sim}(a_t)$ and $\tilde{a}_t^{real}=g_{real}(a_t)$. A \emph{transition gap} changes the traffic dynamics: $T_{sim}(s'\mid s,\tilde{a})\neq T_{real}(s'\mid s,\tilde{a})$. A \emph{reward gap} changes the objective used to train and evaluate the controller: $R_{sim}\neq R_{real}$.

This decomposition defines what the benchmark varies. The observation, action, and transition gaps keep a shared deployment metric, allowing us to measure the performance change between $E_{sim}$ and $E_{real}$. The reward gap changes the metric itself and therefore requires a separate reference: regret against a policy trained directly on the deployment objective.

\subsection{Method Details}
\label{sec:appendix_methods}

\paragraph{Observation.} We evaluate three families: \emph{domain randomization} over the joint corruption distribution; \emph{latent-space domain adaptation}, covering LUSR~\citep{xingDomainAdaptationReinforcement2021}, DARLA~\citep{higginsDARLAImprovingZeroShot2018}, ATC~\citep{stookeDecouplingRepresentationLearning2021}, and a plain VAE embedding~\citep{kingmaAutoEncodingVariationalBayes2022}; and \emph{reconstruction}, an autoencoder that feeds the policy a denoised observation. Domain adaptation here means aligning observation distributions between $E_{sim}$ and $E_{real}$. Domain randomization samples a fresh joint corruption each training episode: detection-zone distance uniform on $[1,100]$\,m, noise $\sigma \in [1.5,5]$, and per-lane failure probability in $[0.1,0.3]$. The latent methods were designed for images; we adapt them to traffic feature vectors by replacing convolutional encoders with multilayer perceptrons, using additive noise and random feature dropout in place of ATC's image crops, and corrupting DARLA's denoising stage by feature masking. In all cases the encoder is pretrained on domain-randomized observations and frozen, and the policy trains on the latent state. The reconstruction baseline is an autoencoder that feeds the policy its denoised output rather than a latent code.

\paragraph{Action (delay).} Against Direct-Transfer we evaluate the \emph{delay-aware prediction} family: Delayed-Q~\citep{dermanActingDelayedEnvironments2023} rolls a forward model through pending actions, and PRLight~\citep{hanMitigatingActionHysteresis2023} predicts post-countdown features in one step. Two ablations isolate what training under the delay buys without prediction: Ignore-Delay (same simulated budget, delay-free training) and Oblivious-Q~\citep{dermanActingDelayedEnvironments2023} (trains under delay but acts on the current observation). Ignore-Delay spends the same simulated budget as the other delay methods but trains in a delay-free simulator and acts as if no delay exists, isolating extra simulated training from delay modeling. Oblivious-Q trains under the delay but selects from the current observation. Delayed-Q learns a forward model and rolls the current state through the queue of pending actions. PRLight predicts post-countdown traffic features in one step with a neighbor-aware module.

\paragraph{Action (phase transitions).} We evaluate domain randomization over random rule tables and grounded action transformation (GAT, UGAT), each with and without deployment-time \emph{action shielding}~\citep{alshiekhSafeReinforcementLearning2018} that masks illegal actions. Running each method with and without the shield separates training-time adaptation from deployment-time masking. Tempe and Bullhead supply NEMA dual-ring tables: an allowed-transition matrix with minimum and maximum dwell times (minimum green and clearance before a switch; force-off at maximum green). Barrier variants disable roughly 43\% of transitions; Cyclic fixes a single successor per phase. Domain randomization trains each episode on a random rule table that disables 20--45\% of allowed transitions while keeping every phase reachable. Action shielding masks the action set to currently legal transitions at decision time and replaces a disallowed action with the highest-ranked allowed one. We run each of domain randomization, GAT, and UGAT both shielded and unshielded, seven deployments per network and setting including Direct-Transfer.

\paragraph{Transition.} We evaluate domain randomization over five car-following parameters~\citep{tobinDomainRandomizationTransferring2017a, mullerBridgingRealityGap2023}, \emph{parameter-space domain adaptation} via simulation-based inference~\citep{ramosBayesSimAdaptiveDomain2019a}, and \emph{grounded action transformation} (GAT~\citep{hannaGroundedActionTransformation2021}, UGAT~\citep{daUncertaintyAwareGroundedAction2023}, JL-GAT~\citep{turnau2025joint} on multi-intersection networks). Domain randomization resamples five car-following parameters (headway time, minimum gap, acceleration, deceleration, vehicle length) each episode from independent clipped Gaussians (Table~\ref{tab:transition_dr}) and uses no real rollouts (300 simulated episodes). Parameter-space domain adaptation fits a neural posterior over the same five parameters, conditioned on transition summary statistics, implemented with the sbi toolkit~\citep{tejero-canteroSbiToolkitSimulationbased2020}. The posterior is retrained as real rollouts accumulate (one per episode), and each training episode draws simulator parameters from the current posterior. Grounding methods (GAT, UGAT, JL-GAT) learn to modify simulated actions so simulated transitions match real ones, using trajectories rather than an explicit parameter model. Domain adaptation and grounding run 100 adaptation episodes, each pairing one simulated and one real rollout with a policy update.

\paragraph{Reward.} We evaluate \emph{reward inference} (recover $w^{real}$ from probe rollouts), \emph{multi-objective RL} (grid of simulator-visible scalarizations, select by real evaluation), and \emph{dynamic reward shaping} (Bayesian weight search scored in $E_{real}$)~\citep{muratoreDataefficientDomainRandomization2021}. Because $R_{real}$ is not computable in $E_{sim}$, $\Delta$ cannot be formed; we score regret against a \emph{reward oracle} trained from scratch in $E_{real}$ on the true objective for 200 real episodes, outside the budget. The oracle separates weak methods from objectives that remain hard even with the true reward. Write $R = \sum_i w_i r_i$. The real objective is $R_{real} = \sum_i w_i^{real} r_i^{real} + \sum_j \alpha_j g_j^{real}$, where $g_j^{real}$ are real-only terms. Each setting is a weighted sum over a shared bank of normalized components (Table~\ref{tab:reward_settings}). Fairness is the spread of cumulative served throughput across approaches, generalizing the two-flow fairness of \citet{raeisDeepReinforcementLearning2021a} to all approaches; CO$_2$ uses the SUMO emission model, and time-to-collision conflicts use a 1.5\,s threshold. Reward Inference deploys short probe policies, each trained on a single reward component, collects feature and return pairs, recovers $w^{real}$ by ridge regression, and retrains in simulation. Multi-objective RL trains over a fixed grid of scalarization weights in simulation (single components, pairwise mixtures, and a uniform mixture) and selects by real evaluation; it covers only simulator-computable objectives. Dynamic Reward Shaping proposes weight vectors with a Gaussian process (expected improvement), briefly fine-tunes the base policy in simulation, and scores each candidate in $E_{real}$. The reward oracle trains for 200 real episodes on the true objective and is scored at the checkpoint of its deterministic real evaluation curve with the highest $R_{real}$; it is an empirical upper reference, not a certified optimum.

\paragraph{Non-RL base controllers.} Fixed-Time and MaxPressure are base controllers like DQN and PressLight, but non-RL: they do not train, no mitigation method applies on top of them, and they deploy directly in $E_{real}$, so they serve as references. We use the Fixed-Time controller from LibSignal~\citep{meiLibsignalOpenLibrary2024}, with a 30-second phase time: it cycles through the phases in order and reads no observations, so the observation gap cannot touch it by construction. MaxPressure~\citep{varaiyaMaxPressureControl2013} switches at every decision to the phase with the highest pressure, the queue difference between upstream and downstream lanes. One run of each covers the DQN and PressLight tables alike, as the two rightmost reference columns of the gap and reward tables in Appendix~\ref{sec:appendix_results}.

\begin{figure}[tbp]
\centering
\includegraphics[width=0.62\textwidth]{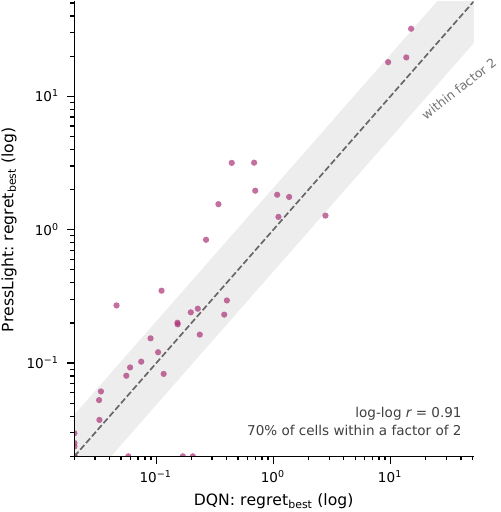}
\caption{\textbf{Consistency across base controllers, reward regret.} Counterpart to \mainfigcontrollers{} for the reward gap: each dot is one network and setting, $\mathrm{regret}_{best}$ under the best budgeted method, DQN on $x$ and PressLight on $y$. The points piled at the lower-left corner are ones where both base controllers reach the oracle. Band and $r$ as in \mainfigcontrollers{}.}
\label{fig:controllers_reward}
\end{figure}

\begin{figure}[tbp]
\centering
\includegraphics[width=\textwidth]{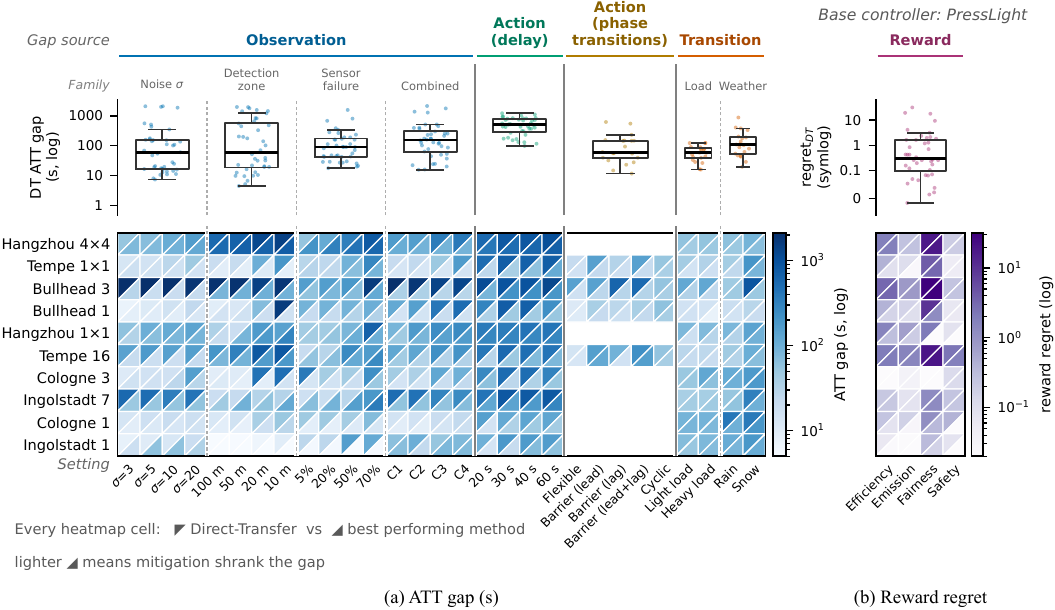}
\caption{\textbf{The Sim-to-Real gap by setting family and by network for PressLight.} Companion to \mainfignetworks{}: identical construction, same row order, every value is a single run of the PressLight base controller.}
\label{fig:networks_presslight}
\end{figure}

\subsection{Gap Design}
\label{sec:appendix_gap_design}

This subsection records where each induced gap comes from: the real-world evidence for it and the sources of the setting values. The values themselves are listed in the tables above and in the main text.

\paragraph{Observation gap.} Field sensing degrades in three ways: readings are noisy~\citep{bachechiDetectionClassificationSensor2022, rodriguesRobustDeepReinforcement2019}, sensors fail outright, and every sensor covers a limited detection zone. The noise settings use the zero-mean Gaussian sensor model of \citet{aslaniAdaptiveTrafficSignal2017}; we keep their model and widen the severity range to $\sigma \in \{3, 5, 10, 20\}$. The sensor-failure settings follow field studies. In a seven-month study of 2{,}263 freeway loop detectors, 25\% reported no data at all and the rest were missing 29\% of their measurements on average~\citep{hutchinsProbabilisticAnalysisLargeScale2010}, and stuck-at-zero output is a standard anomaly class~\citep{bachechiDetectionClassificationSensor2022}. We therefore model failure as stuck-at-zero: each lane detector fails independently with probability $p \in \{0.05, 0.2, 0.5, 0.7\}$, drawn once per episode and held, so a failed detector stays failed. The lower rates bracket the observed failure shares; the upper ones are stress settings. The detection-zone distances of 100, 50, 20, and 10\,m span the sensor spectrum of the traffic detector handbooks~\citep{kleinTrafficDetectorHandbook2006, kleinTrafficDetectorHandbook2006a}: video image processors reach roughly 60\,m, loop installations cover a few meters up to about 35\,m, and only high-speed dilemma-zone designs approach 100\,m. The combined settings of Table~\ref{tab:obs_combine} apply all three corruption sources at once, at severities that could occur together in the field.

\paragraph{Action gap (delay).} Real controllers do not switch instantly: many deployments display a countdown, and a requested change takes effect only after it elapses~\citep{panImpactCountdownSignals2023}. We model the countdown as a constant action delay~\citep{dermanActingDelayedEnvironments2023}. Decisions are taken every 10 seconds and $E_{real}$ applies delays of 20, 30, 40, and 60 seconds, so two to six decisions are in flight at any time. The mitigation methods assume a fixed 20-second countdown, so the larger delays also measure how a method degrades when the real countdown exceeds the assumed one.

\paragraph{Action gap (phase transitions).} Real signal programs forbid many transitions through minimum green times, clearance intervals, and fixed sequences~\citep{NEMAStandardsPublication2021, urbanikSignalTimingManual2015}. The Tempe and Bullhead networks ship with their real NEMA dual-ring tables, and the five settings are derived from them: Flexible keeps the dwell-time rules but forbids no transition, the three Barrier variants disable roughly 43\% of transitions, and Cyclic fixes a single successor per phase. Only these two locations carry real signal plans, which is why this gap is evaluated on four networks rather than ten.

\paragraph{Transition gap.} Traffic dynamics differ between simulation and reality through car-following behavior, demand, and weather~\citep{harthAutomatedCalibrationTraffic2021, rodriguesRobustDeepReinforcement2019}. The simulator pair is the first source: CityFlow is $E_{sim}$ and SUMO is $E_{real}$, and their dynamics never match exactly, which is the no-gap residual of Table~\ref{tab:calibration}. On top of it we perturb the Krauss car-following parameters~\citep{kraussMicroscopicModelingTraffic1998} of $E_{real}$. The four settings of Table~\ref{tab:transition_settings} take their values from the perturbations of \citet{daUncertaintyAwareGroundedAction2023}, in line with published car-following calibration studies~\citep{pourabdollahCalibrationEvaluationCar2017}: the two loading settings raise the startup delay, modeling loaded vehicles as slower to launch, and the two weather settings cut acceleration and braking, mildly for rain and severely for snow. The domain randomization distributions of Table~\ref{tab:transition_dr} combine the randomization of \citet{mullerBridgingRealityGap2023} with the calibrated parameter values of \citet{schraderCalibratingCarFollowingModels2024}.

\paragraph{Reward gap.} Traffic agencies weigh delay against safety, emissions, noise, and pedestrian considerations~\citep{urbanikSignalTimingManual2015}, while common simulators expose only a subset of these quantities. The four settings of Table~\ref{tab:reward_settings} cover the two halves of the gap: Efficiency-aligned shares its components with the training reward and acts as a sanity check, and the three heavy settings each put the dominant weight on a term the simulator-trained policy cannot observe. One design correction is worth recording. The fairness objective originally weighted only queue and the fairness spread, and that was exploitable: the spread of served throughput is near zero under gridlock, because equal starvation is equal. A gridlocked policy scored better than an honestly fair one. The objective therefore also charges for starvation through the waiting component, which is orders of magnitude larger under gridlock, so the exploit can no longer win. The fairness measure itself is unchanged.

\subsection{Reward Metrics}
\label{sec:appendix_reward_metrics}

The hidden objective $R_{real} = -\sum_i w_i\, \phi_i$ sums normalized cost components $\phi_i$ with the weights of Table~\ref{tab:reward_settings}. Each raw component is divided by a fixed per-component normalizer before weighting, so the weights express relative importance rather than unit conversions. Queue, delay, and waiting are the standard traffic quantities; this subsection defines the components that are not, and how each is computed.

\paragraph{Emission and fuel.} CO$_2$ and fuel come from SUMO's HBEFA-based emission model~\citep{SUMO2018}, read per lane and accumulated over the episode; the tables report episode totals in kilograms. CityFlow has no emission model, so both are computable only in $E_{real}$, and a simulator-trained policy can reach them only through correlated proxies.

\paragraph{Safety conflicts.} Physical safety is measured by surrogate-safety conflicts, because SUMO traffic is collision-free by construction: junction conflicts resolve as emergency stops, not crashes, so near-crashes carry the signal that crashes cannot. A vehicle enters a conflict when its car-following time to collision drops below a 1.5\,s screening threshold. For a follower closing on its leader, $\mathrm{TTC} = d / (v_{f} - v_{l})$, where $d$ is the bumper-to-bumper distance to the leader, taken from SUMO's leader lookup. Both the measure and the threshold follow the surrogate-safety review of \citet{mahmudApplicationProximalSurrogate2017}, the reference SUMO's own SSM documentation uses. Each vehicle is counted at most once per episode, and the conflict is attributed to the intersection controlling the vehicle's lane. We compute this per step from the leader relation rather than through SUMO's SSM device: the device buffers every open encounter, its memory grows with steps times vehicle pairs, and it exhausts memory on long congested episodes. The per-step computation reports the same TTC quantity the device reports for car-following situations. Its scope is rear-end conflicts, the mode signal control actually induces; crossing and merging geometries are not evaluated.

\paragraph{Fairness.} A controller is unfair when it keeps serving one approach while another with waiting vehicles is starved. Let $C_i(t)$ be the served throughput of approach $i$: the number of vehicles that have left its incoming lanes up to step $t$. An approach is demand-active when it has waiting vehicles, $w_i(t) > 0$. The fairness cost is the largest pairwise throughput gap between demand-active approaches,
\[
\phi_{fair}(t) = \max_{i,\,j} \,\bigl| C_i(t) - C_j(t) \bigr|,
\]
where $i$ and $j$ range over the demand-active approaches, with all approaches weighted equally. The cost is zero when fewer than two approaches are demand-active, since a controller cannot be unfair to a single approach. This generalizes the two-flow fairness of \citet{raeisDeepReinforcementLearning2021a} to all approaches of an intersection, and it is computable in both simulators.

\paragraph{Logged but unweighted.} Fuel, emergency stops, and collisions are measured and logged for every reward-gap run but carry no weight in any deployment objective. Collisions carry almost no signal, since SUMO resolves would-be crashes as emergency stops, and are kept at weight zero. The component bank also carries pressure and phase-switch-rate components; no deployment objective weights them.

\section{Full Results}
\label{sec:appendix_results}


Two kinds of tables follow. \emph{Gap tables} report the per-setting ATT gap $\Delta$ (mean real ATT minus the calibrated $E_{sim}$ ATT) for every network, method, and base controller, as mean $\pm$ std across the top five evaluation checkpoints of the protocol; Direct-Transfer is a single deterministic evaluation and has no deviation. The reward gap tables report the hidden objective $R_{real}$ instead, and the phase-transition gap tables are followed by a table of attempted disallowed transitions and force-offs per episode. \emph{Core-metric tables} report the corresponding absolute values in $E_{real}$: ATT, average queue length, average delay, and throughput per run. Average delay appears only in absolute form because its gap is not comparable across the two simulators. The reward-gap metric tables further report the hidden components in their native units: total CO$_2$ over the episode (kg), fairness as the spread of served throughput across approaches (vehicles per episode), and TTC conflicts per episode. These native units are the ones to use for cross-network severity, since $R_{real}$ itself is not comparable across networks. The reward tables carry the oracle as a reference row, and the phase-transition tables report each adaptation method twice, with and without the deployment shield. The gap and reward tables also carry the two non-RL base controllers in their rightmost columns, Fixed-Time and MaxPressure, deployed directly in $E_{real}$ as references. They do not train, so one run of each covers the DQN and PressLight tables alike. Their gap is measured against their own run in $E_{sim}$, shown together with the absolute real ATT in brackets. Unlike the pretrained checkpoints, these reference runs are not calibrated across the two simulators, so a large negative reference gap means the simulators disagree under congestion, not that the induced gap helps; the bracketed ATT is the value to compare against the RL base controllers.

\input{tables/full_results}

\end{document}